\documentclass[10pt,twocolumn,letterpaper]{article}

\usepackage{times}
\usepackage{epsfig}
\usepackage[dvipsnames]{xcolor} 
\usepackage{graphicx} 
\usepackage{amsmath}
\usepackage{amssymb}
\usepackage{multirow}
\usepackage{gensymb}
\usepackage{soul}
\usepackage{colortbl}
\usepackage{color}
\usepackage[pagenumbers]{iccv} 

\definecolor{iccvblue}{rgb}{0.21,0.49,0.74}

\newcommand{\PAR}[1]{\vskip4pt \noindent{\bf #1~}}
\newcommand{\graycell}{\cellcolor{gray!15}}

\newlength\savewidth
\usepackage[pagebackref,breaklinks,colorlinks,allcolors=iccvblue]{hyperref}
\newcommand\blfootnote[1]{%
  \begingroup
  \renewcommand\thefootnote{}\footnote{#1}%
  \addtocounter{footnote}{-1}%
  \endgroup
}

\def\paperID{9678} 
\def\confName{ICCV}
\def\confYear{2025}

\title{Mind the Gap: Aligning Vision Foundation Models to Image Feature Matching}

\author{
    \quad Yuhan Liu 
    \quad Jingwen Fu
    \quad Yang Wu 
    \quad Kangyi Wu
    \quad Pengna Li
    \\
    \quad Jiayi Wu
    \quad Sanping Zhou
    \quad Jingmin Xin$^{\dagger}$
    \\
             National Key Laboratory of Human-Machine Hybrid Augmented Intelligence,\\
             National Engineering Research Center for Visual Information and Applications,\\
             Institute of Artificial Intelligence and Robotics, Xi’an Jiaotong University\\
             {\tt\small \{liuyuhan200095, wuyang\_cc, wukangyi747600, sauerfisch\}@stu.xjtu.edu.cn, jwfu99@gmail.com}\\
             {\tt\small wujiayi0101@163.com, spzhou@xjtu.edu.cn, jxin@mail.xjtu.edu.cn}
            \quad 
}

\begin{document}
\maketitle
\blfootnote{$^\dagger$Corresponding author.}

\begin{abstract}
Leveraging the vision foundation models has emerged as a mainstream paradigm that improves the performance of image feature matching.
However, previous works have ignored the \textbf{misalignment} when introducing the foundation models into feature matching. 
The misalignment arises from the discrepancy between the foundation models focusing on single-image understanding and the cross-image understanding requirement of feature matching. 
Specifically, 1) the embeddings derived from commonly used foundation models exhibit discrepancies with the optimal embeddings required for feature matching; 2) lacking an effective mechanism to leverage the single-image understanding ability into cross-image understanding.
A significant consequence of the misalignment is they struggle when addressing multi-instance feature matching problems. 
To address this, we introduce a simple but effective framework, called IMD (Image feature Matching with a pre-trained Diffusion model) with two parts: 
1) Unlike the dominant solutions employing contrastive-learning based foundation models that emphasize global semantics, we integrate the generative-based diffusion models to effectively capture instance-level details. 
2) We leverage the prompt mechanism in generative model as a natural tunnel, propose a novel cross-image interaction prompting module to facilitate bidirectional information interaction between image pairs.  
To more accurately measure the misalignment, we propose a new benchmark called IMIM, which focuses on multi-instance scenarios. 
Our proposed IMD establishes a new state-of-the-art in commonly evaluated benchmarks, and the superior improvement 12\% in IMIM indicates our method efficiently mitigates the misalignment.

\end{abstract}

\section{Introduction}

\begin{figure}[tp]
    \centering
    \includegraphics[width=1\linewidth]{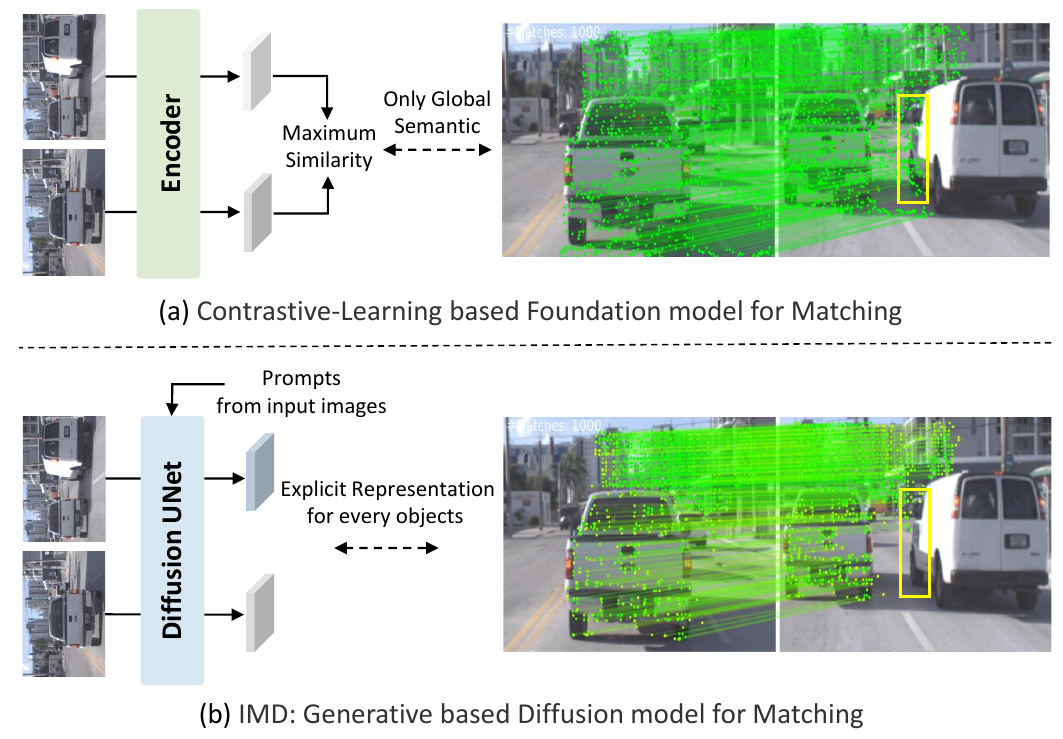}
    \vspace{-0.6 cm}
    \caption{\textbf{The main idea of our proposed IMD.}
    Previous works \cite{edstedt2024roma,liu2024semantic} failed in multi-instance scenarios due to the misalignment.  Our IMD leverages the explicit representation of specific instances within the diffusion model with an interaction module, accurately matching the specific instance.     
     }
    \vspace{-0.4 cm}
    \label{fig-1}
\end{figure}

Image feature matching serves as the foundation for many computer vision tasks, finds extensive applications in areas as image stitching~\cite{liao2023recrecnet,nie2021unsupervised}, 3D reconstruction \cite{schonberger2016structure,schonberger2016pixelwise} and visual localization \cite{cai2024voloc,sarlin2019coarse,sarlin2021back}. 
Over the past decade, image feature matching has evolved from hand-crafted approaches \cite{bay2006surf, rublee2011orb} to learning-based methods \cite{detone2018superpoint,potje2023enhancing,revaud2019r2d2}. Recently, numerous learnable image matchers \cite{edstedt2023dkm,sun2021loftr,tang2022quadtree,lindenberger2023lightglue} trained on real-world datasets \cite{dai2017scannet,li2018megadepth} have been proposed achieving ever-improving performance. 
It has been observed that such methods tend to overfit to a specific training distribution or domain and
fail to generalize on unseen data fail to generalize on unseen data \cite{jiang2024omniglue}.
Recent approaches \cite{edstedt2024roma,jiang2024omniglue,cao2023improving} have demonstrated that integrating vision foundation models pre-trained on large-scale datasets can effectively solve this issue, achieving superior generalization ability. 
However, our investigation identifies a non-negligible misalignment when adapting these vision foundational models to image feature matching.

This misalignment comes from the discrepancy between the vision foundation models focus on understanding the content of the single-image and the cross-image relational understanding requirements for feature matching. 
The consequence of the misalignment is the poor performance of these methods in multi-instance matching problems as illustrated in Figure \ref{fig-1}.  
In this case, multiple instances of the same object class exist, which impose high requirements on cross-image understanding to locate the correct instance.
To address the misalignment issue, this study identifies two critical challenges that need to be confronted:

\textbf{Challenge I:} \quad 
\textit{Which properties that the foundation model has to equip in order to reduce the misalignment?} Previous works leverage contrastive-learning based foundation model \cite{edstedt2024roma,jiang2024omniglue,liu2024semantic}, which maximum the similarity between features prioritize global semantics. 
However, as illustrated in Figure \ref{fig-1}(a), these foundation models are suboptimal for image feature matching, such contrastive objectives often lose information about specific instances and objects \cite{zhang2024tale}, which are crucial for accurate matching.
To solve this issue, we argue that generative-based foundation models serve as a more suitable alternative.
The reason is that the feature of generative-based models contain various objects and instances each with distinctive appearances and structures \cite{tang2023emergent,samuel2024s}.
As demonstrated in Figure \ref{fig-1}(b), this information is crucial for image feature matching, particularly in multiple instances scenarios.

\textbf{Challenge II:}\quad \textit{How to design a cross-image interaction mechanism aligning with the matching tasks?} Although the foundation models have strong single-image understanding ability, they can not lead to cross-image understanding by nature. This paper identifies the condition mechanism of the generative-based model as a natural tunnel for the cross-image information interaction. Inspired by this, we intend to utilize each image as the condition for the other image.

In this paper, we propose {\textbf{\textit{IMD}}}: Image feature Matching with Diffusion models, seamlessly adapt foundation models to matching tasks. The IMD contains two parts:
\textbf{1)} We introduce a novel feature extraction paradigm based on the phenomenal generative-based diffusion models, exploring its ability of self-image understanding while moving toward cross-image understanding.
Different from previous works utilizing the progressive denoising process \cite{zhang2024diffglue,nam2023diffusion}, we employ the frozen UNet decoder as a backbone to directly process clean natural images to obtain the explicit representation for every instance. 
\textbf{2)} To facilitate interaction between image pairs, we prompt the extraction of diffusion backbone with a novel Cross-image Interaction Prompt Module (CIPM).  
Specifically, we design a personalized prompt for each image that was conditioned by the image pair to enhance relevance and discriminability.
We also propose to utilize the cross-attention layers to proceed with explicit interaction and comparison between the image pair.
CIPM enforces the model to capture the underlying geometric and semantic relationships between images during the feature extraction process itself, resulting in more correlated and comprehensive image features.

For a more comprehensive analysis, we construct a new benchmark IMIM based on a recently published video tracking and segmentation dataset \cite{athar2023burst}. We observe that existing benchmarks predominantly comprise images containing either a single, prominent object, enabling semantic-based approaches to attain high performance. In contrast,
our IMIM contains images with multiple instances of the same semantic (e.g., two cars running), which imposes more stringent requirements for preserving instance-level details. We report a significant improvement of 12\% in matching accuracy, demonstrating that our method successfully alleviates the misalignment.

To summarize, we make the following contributions:
\begin{itemize}
    \item 
    We identify the misalignment when introducing the foundation models into the feature matching tasks with two challenges for solving the misalignment.   

    \item We propose IMD to achieve seamless alignment with 1) a novel feature extraction pipeline for matching tasks and 2) cross-image interaction prompt module towards cross-image understanding.

    \item IMD is demonstrated to achieve state-of-the-art results across a wide range of benchmarks. What's more, a new dataset is introduced to evaluate the performance of models in the multi-instance situation.
\end{itemize}

\section{Related Work}
\paragraph{Local Feature Matching.}
Feature matching methods can be broadly classified into three categories: sparse, semi-dense, and dense. 
Traditional sparse feature matching has been tackled through keypoint detection and description, followed by matching the descriptors \cite{lowe2004distinctive,bay2008speeded,rublee2011orb,barroso2019key,detone2018superpoint,sarlin2020superglue}.
More recently, the semi-dense approach ~\cite{sun2021loftr,chen2022aspanformer,liu2024semantic,tang2022quadtree} has emerged, also known as detector-free methods, replacing keypoint detection with dense matching at a coarse scale, followed by mutual nearest neighbor extraction and subsequent refinement.
The dense approach ~\cite{edstedt2023dkm,edstedt2024roma,zhu2023pmatch} focuses on estimating a dense warp, aiming to identify every potential pixel-level correspondence. However, they are typically much slower than sparse and semi-dense approaches. 
\begin{figure*}[tp]
    \centering
    \includegraphics[width=1\linewidth]{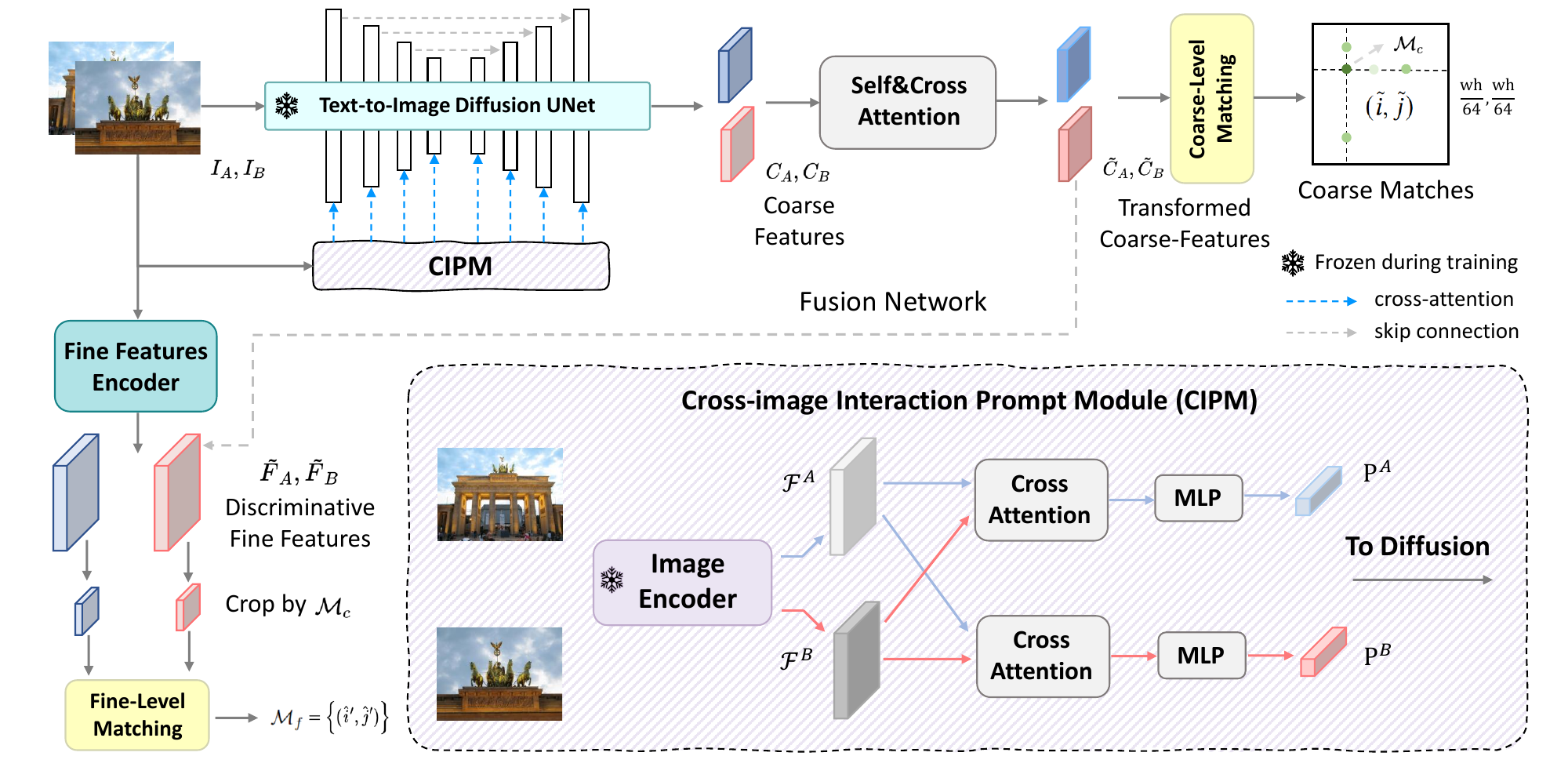}
    \vspace{-0.6 cm}
    \caption{\textbf{Pipeline Overview.} Given an image pair, we first encode them with the image encoder and employ the cross-attention mechanism to proceed with explicit interaction between images. With the images and its prompt containing interactive information as input, we extract their coarse features $C_{A}, C_{B}$ from a frozen text-to-image diffusion UNet. Then, we refine the coarse features for more discriminative by iteratively applying self- and cross-attention mechanisms. The coarse-level matching generates pixel-to-pixel matches $\mathcal{M}_{c}$ at 1/8 scale. To refine coarse matches, discriminative fine features $\tilde{F}_{A}, \tilde{F}_{B}$ are obtained by fusing the transformed coarse features with original fine features from the specialized fine feature encoder. Feature patches are then cropped, centered around each coarse match $\mathcal{M}_{c}$. The fine-level matching is followed to obtain sub-pixel matches $\mathcal{M}_{f}$.  
     }
    \vspace{-0.3 cm}
    \label{fig-2}
\end{figure*}


\PAR{Vision Foundation Models for Matching.}
There exist prior works leveraging the vision foundation models to improve matching performance. MESA \cite{zhang2024mesa} introduced the use of SAM \cite{kirillov2023segment} for area matching, narrowing the search space. Nevertheless, the coarse-grained utilization fails to fully leverage the knowledge of foundation models.
Another series of works \cite{cao2023improving,jiang2024omniglue,edstedt2024roma,liu2024semantic} employees vision foundation models as feature extractor directly, significantly improving the performance. 
However, we identify that they ignored the critical misalignment between foundation models and the matching tasks.  
Specifically, the commonly used DINOv2 \cite{oquab2023dinov2} often loses information about specific instances. Meanwhile these models lack information interaction between image pairs, leading to failures in multiple instances scenes.
In contrast, we propose to integrating the generative-based diffusion models to address the misalignment   
with a novel cross-image interaction module, achieving excellent performance in multi-instance scenes.

Building on superior generative performance of diffusion models, recent studies have started investigating the internal representations within diffusion models \cite{samuel2024s,zhang2024telling,zhao2023unleashing,xu2023open}.
DIFT \cite{tang2023emergent} and SD+DINO \cite{zhang2024tale} 
demonstrated that extracting features from denoising module enables semantic matching between two objects.
In addition, DIFT has also shown that the image features from frozen diffusion models can act as powerful descriptors for feature matching tasks. 
Our work is closely supported by these efforts, but we instead 
explore the misalignment when adapting the foundation models into feature matching and how to overcome.

\PAR{Cross-image Interaction.}  
Mainstream approaches such as RoMa \cite{edstedt2024roma} and OmniGlue \cite{jiang2024omniglue} utilize visual foundation models to process individual images, thereby obtaining independent image features. The relationship between the images is obtained through a subsequent attention mechanism. However, we argue that performing matching using image features extracted in this way limited performance through the image features lacking correlation. CroCo \cite{weinzaepfel2022croco} and CroCov2 \cite{weinzaepfel2023croco} leverage cross-view completion as a pre-training objective to inherently encourage the model to learn correlated features cross images.
SD4Match \cite{li2024sd4match} has underscored the importance of the relevance of cross-image in semantic matching,
which employ a shared prompt conditioned on the image pairs for diffusion model. 
Nevertheless, this work faces notable issue: 
this shared prompt introduces correlation while leading to worse discriminative. 
In contrast, our proposed CIPM provided the personalized prompts for each image, guiding feature extraction to increase relevance while maintaining discriminancy.
Our IMD not only improves information transfer across images but also results in correlated features.

\section{Method}

In this section, we present IMD, a new detector-free local feature matching method to overcome the misalignment between vision foundation models and feature matching.  
IMD adopts a typical two-stage framework, starting with coarse-level correspondences and subsequently refining them to sub-pixel positions.
Given a pair of images $I_{A}$ and $I_{B}$, the UNet component within diffusion models generates coarse feature maps ($1/8$) from input images based on the prompt produced by our designed prompting module (Sec. \ref{sec3.2}). Specifically, we propose the cross-image interaction prompt module to explicitly facilitate cross-image interaction. With these extracted features, the self- and cross-attention based coarse matching is performed in coarse-level features (Sec. \ref{sec3.3}). Using the specialized fine features encoder, we extract the fine feature maps ($1/2$) from image pairs. 
According to the coarse matches, a subpixel-level refinement module is employed to predict fine-level matches (Sec \ref{sec3.4}). 
An overview of our approach is illustrated in Figure \ref{fig-2}.

\subsection{Text-to-Image Diffusion Model}
\label{sec3.1}

We begin with a brief overview of the diffusion model and its adaptation for feature extraction.

\PAR{Background.}
Diffusion models \cite{ho2020denoising,song2020denoising, DynamicID} are generative models that are designed to transform a normal distribution to an arbitrary data distribution. This model involves two processes: forward and reverse. In the forward process, Gaussian noise of varying magnitudes is added to clean data points to create noisy data samples. A clean image \(I_{0}\) is gradually degraded to a noisy image \(I_{t}\) through a series of Gaussian noise additions, following the equation:
\begin{equation}
    I_{t}=\sqrt{\alpha _{t} } I _{0}+(\sqrt{1-\alpha _{t}} ) \epsilon \label{eq1},
\end{equation}
where $\epsilon \sim \mathcal{N}(0, 1)$ denotes the randomly sampled noise, and $t \in [0, T]$ represents the ``time'' in the diffusion process with larger time steps corresponding to higher levels of noise. The noise amount is controlled by the pre-defined noise schedule $\{\alpha _t \}^T_t$.
In the reverse process, a neural network \( f_\theta \) is trained to take \( I_t \) and the time step \( t \) as inputs to predict the added noise \( \epsilon \). For image generation, \( f_\theta \) is typically parametrized as a UNet \cite{rombach2022high,dhariwal2021diffusion}. Once trained, \( f_\theta \) can be utilized to ``reverse" the diffusion process. Thus, by sampling from a normal distribution, the corresponding ``original image'' can be reconstructed through iterative removal of the noise \( \epsilon \).

\PAR{Diffusion Model as Feature Extractor.}
Our framework is built upon the widely used Stable Diffusion (SD) \cite{rombach2022high} models, which perform the denoising process in a learned latent space using a UNet architecture. SD also exhibits impressive capability to generate high-quality images conditioned on input prompts, such as text or images. 
Given an input image \(I_{0}\) and a specific timestep \(t\), \(I_{0}\) is first encoded into its latent representation \(z_{0}\) using a VAE encoder $\mathcal{E}$, which is subsequently corrupted to \(z_t\) using Eq. (\ref{eq1}). 
During this noise prediction process, with the latent code \(z_t\) at timestep \(t\) and the prompt embedding \(P\) as inputs, we extract the feature map \(F\) from the output of an intermediate layer of denoising UNet. This process can be formally expressed as follows:
\begin{equation}
\begin{aligned}
    z_0 &= \mathcal{E}(I_0), \\
    z_t &= \sqrt{\alpha_t}z_0 + \sqrt{1 - \alpha_t}\epsilon, \\
    F &= \text{UNet}(z_t, t, P).
    \label{eq2}
\end{aligned}
\end{equation}

Another important consideration is the selection of the timestep \(t\) and the network layer from which features are extracted. Studies indicate larger \(t\) and earlier network layers tend to produce more semantically-aware features, while smaller \(t\) and later layers emphasize low-level details \cite{tang2023emergent}.
For a detailed overview of the hyper-parameter values utilized in this study, please refer to the Suppl. For simplicity, we omit the VAE encoding step and directly refer to the input to the UNet as the image \(I\) rather than its latent representation \(z\) in the following sections.

\subsection{Cross-image Interaction Prompt Module}
\label{sec3.2}

Text-to-image diffusion models pre-trained on extensive image-text pair datasets, exhibit remarkable controllability through customizable prompt mechanisms. 
During each step of the denoising process, diffusion models perform cross-attention between the text embedding of prompts and UNet features leverage the text input to guide the denoising direction of the noisy image. 
As shown in Eq.(\ref{eq2}), the diffusion model's visual representation \(F\) of image \(I\) relies on its associated prompt \(P\). 
This motivated us to explore the possibility of using another image as a prompt to generate interactive image features.

Recent studies have explored generating the prompts with input images for tasks such as semantic correspondence \cite{li2024sd4match,zhang2024tale} and image segmentation \cite{xu2023open}.
In this context, the prompt is generated from the input image itself in single-image task or a shared prompt by concatenating the features from the image pair in tasks involving a pair of images. Although this design has proven effective for the aforementioned tasks, it may be not the optimal choice for facilitating image interaction in feature matching. Our reasons are:
\begin{enumerate}[leftmargin=*,noitemsep, topsep=0pt]
    \item Feature matching requires the model to focus on the local features of the images. Utilizing a shared prompt for feature extraction across image pairs can result in the loss of detailed information from the images.
    \item Simply concatenating image features is insufficient to fully capture the relationships between them, as it lacks explicit learning of dependencies and interactive information between the images.
\end{enumerate}
We therefore propose a novel cross-image interaction prompt module (CIPM) and illustrate its architecture in the bottom of Figure \ref{fig-2}. Given a pair of clean images $I_A$ and $I_B$, we leverage a pre-trained frozen image encoder $\mathcal{V}$, e.g., from CLIP \cite{radford2021learning} to extract the image features $\mathcal{F}^A$ and $\mathcal{F}^B$. Then we adopt three $1\times1$ convolution layers to encode the input image feature $\mathcal{F}^A$ and $\mathcal{F}^B$ into features $\phi_{Q,I}$, $\phi_{K,I}$, $\phi_{V,I}$. We apply the cross-attention mechanism between features from input images $I_A$ and $I_B$.We further utilize a learned MLP to transform the image embedding into the prompt embedding $P_A$ and $P_B$, which is then fed into the text-to-image diffusion UNet, denoted as:
\begin{equation}
\begin{gathered}
    {P}_{A}=\text{MLP}\circ Softmax( \phi_{Q,I_A}\phi_{K,I_B}/\sqrt{d_{k} }) \phi_{V,I_B},\\
    {P}_{B}=\text{MLP}\circ Softmax( \phi_{Q,I_B}\phi_{K,I_A}/\sqrt{d_{k} }) \phi_{V,I_A},
\end{gathered}
\end{equation}
where $d_{k}$ denote the scale factor. Finally, the UNet of the text-to-image diffusion model, combined with the CIPM, constitutes IMD's feature extractor, which generates the visual representation for the given image pairs. Formally, the coarse features $C_A$ and $C_B$ are computed as:
\begin{equation}
\begin{gathered}
    C_{A}=\text{UNet}(I_A, t, {P}_{A}),\\ 
    C_{B}=\text{UNet}(I_B, t, {P}_{B}).
\end{gathered}
\end{equation}

The attention mechanism facilitates interaction between the input images $I_A$ and $I_B$ within the feature space, thereby capturing correspondence information more explicitly compared to merely concatenating the image pair channel-wise before feeding them into the network \cite{li2024sd4match}. Furthermore, the transformer-based network demonstrates greater robustness when the two input images have significant differences.

\subsection{Coarse-Level Matching Module}
\label{sec3.3}

We also conduct the attention mechanisms for coarse features to improve discriminativeness and compute coarse-level matches using the transformed coarse feature maps \(\tilde{C}_A\) and \(\tilde{C}_B\). These coarse correspondences provide rough matching regions that are refined in the subsequent fine-level refinement module. 
The score matrix \(\mathcal{S}\) is generated by computing the pairwise cosine similarity between coarse-level features \(\tilde{C}_A\) and \(\tilde{C}_B\). 
We also apply the dual-softmax of \(\mathcal{S}\) to obtain the probability of mutual nearest matching, following ~\cite{cao2023improving,chen2022aspanformer, sun2021loftr}. The coarse matches \( \{\mathcal{M}_c\} \) are determined by selecting matches that exceed a predefined score threshold \(\tau\) and satisfy the mutual nearest neighbor (MNN) constraint.

\subsection{Subpixel-Level Refinement Module}
\label{sec3.4}
As illustrated in Figure \ref{fig-2}, with the established coarse matches \( \{\mathcal{M}_c\} \), we refine these matches to achieve sub-pixel accuracy following \cite{wang2024efficient}. This module consists of a specialized fine feature encoder for generating discriminative fine features, followed by the fine-level refinement to produce the final matches \( \{\mathcal{M}_f\} \).

\paragraph{Fine Features Encoder.} 
In prior works \cite{chen2022aspanformer,sun2021loftr}, the FPN \cite{lin2017feature} is employed for producing a feature pyramid of coarse and fine features used for coarse matching and later refined to a sub-pixel level. 
This is problematic when using SD features as only coarse features at 1/8 dimension of the original image exist. Hence, we utilize a specialized ConvNet obtained fine features at 1/2 dimension. For efficiency, these fine image features are fusion with the previously transformed coarse features to obtain discriminative fine features $\tilde{F}_{A}$, $\tilde{F}_{B}$ in the original image resolution eliminating the additional transform networks \cite{sun2021loftr}. Then local feature patches are cropped on fine feature maps centered at each coarse match \( \{\mathcal{M}_c\} \).

\paragraph{Fine-Level Matching.}
To refine the coarse matches, we first compute the local patch score matrix $\mathcal{S}_l$ with the fine feature patches. MNN searching is applied on $\mathcal{S}_l$ to get intermediate pixel-level fine matches. We select the top-1 fine match for each coarse match by ranking the correlation scores. Then the feature of each point in ${I}_A$ is correlated with a $3\times3$ feature patch centered around its fine match in ${I}_B$. The softmax operation is applied to obtain a match distribution matrix, and the final refined match is determined by calculating the expectation.

\begin{table*}[t]
\caption{\textbf{Results of multi-instance evaluation on IMIM and two-view pose estimation on the MegaDepth \cite{li2018megadepth}, ScanNet \cite{dai2017scannet} datasets.} Methods are grouped into 3 groups: 1) methods that are zero-shot and not fine-tuned on the training data, 2) sparse methods, and 3) semi-dense methods. All the IMD results have \graycell gray background for easy lookup and annotate \textbf{best} results. The extended version including dense methods, is provided in the Suppl.}
\vspace{-0.1cm}
\centering
\resizebox{0.95\linewidth}{!}{ 
\setlength\tabcolsep{4pt} 
\begin{tabular}{@{}clccccccc@{}}
\toprule
\multirow{2}{*}{\textbf{Category}} & \multirow{2}{*}{\textbf{Method}} & \multicolumn{3}{c}{\textbf{MegaDepth}} & \multicolumn{3}{c}{\textbf{ScanNet}} & \multirow{2}{*}{\textbf{IMIM (\%)}} \\ \cmidrule(l){3-5}\cmidrule(lr){6-8}
    
                          &                         & AUC@5\degree & AUC@10\degree & AUC@20\degree & AUC@5\degree & AUC@10\degree & AUC@20\degree & \\ \midrule
\multirow{3}{*}{Zero-Shot} 
                               & CLIP \cite{radford2021learning}~\tiny{ICML'21} & 30.8 & 48.1 & 63.2 & 10.1 & 20.6 & 31.3 & 54.4 \\
                               & DINOv2 \cite{oquab2023dinov2}~\tiny{Arxiv'23} & 32.5 & 50.8 & 65.3 & 13.0 & 28.5 & 40.8 & 57.9 \\ 
                               & DIFT \cite{tang2023emergent}~\tiny{NeurIPS'23} & 38.4 & 55.9 & 70.5 & 15.7 & 32.0 & 45.1 & 61.2 \\
\midrule\midrule
\multirow{3}{*}{Sparse} & SP \cite{detone2018superpoint}+NN~\tiny{CVPRW'23} & 31.7 & 46.8 & 60.1 & 7.5 & 18.6 & 32.1 & 55.9 \\
                             & OmniGlue \cite{jiang2024omniglue}~\tiny{CVPR'24} & 47.4 & 65.0 & 77.8 & \textbf{31.3} & \textbf{50.2} & \textbf{65.0} & \textbf{77.6} \\ 
                             & SP \cite{detone2018superpoint}+LG \cite{lindenberger2023lightglue}~\tiny{ICCV'23} & \textbf{49.9} & \textbf{67.0} & \textbf{80.1} & 14.8 & 30.8 & 47.5 & 60.5 \\
\midrule
\multirow{10}{*}{Semi-Dense} 

& LoFTR \cite{sun2021loftr}~\tiny{CVPR'21} & 52.8 & 69.2 & 81.2 & 16.9 & 33.6 & 50.6 & 68.9 \\
& RCM \cite{lu2024raising}~\tiny{ECCV'24} & 53.2 & 69.4 & 81.5 & 17.3 & 34.6 & 52.1 & - \\
                                & Effcient LoFTR \cite{wang2024efficient}~\tiny{CVPR'24} & 56.4 & 72.2 & 83.5 & 19.2 & 37.0 & 53.6 & 70.6 \\
                                & EcoMatcher \cite{chen2024ecomatcher}~\tiny{ECCV'24} & 56.5 & 72.0 & 83.4 & - & - & - & - \\
                                & HomoMatcher \cite{wang2024homomatcher}~\tiny{AAAI'25} & 57.8 & 73.5 & 84.4 & 22.1 & 40.9 & 57.5 & - \\
                                & TopicFM \cite{giang2023topicfm}~\tiny{AAAI'23} & 58.2 & 72.8 & 83.2 & 17.3 & 34.5 & 50.9 & 76.5 \\
                                & MESA\_ASpan \cite{zhang2024mesa}~\tiny{CVPR'24} & 58.4 & 74.1 & 84.8 & - & - & - & - \\
                                & CasMTR \cite{cao2023improving}~\tiny{ICCV'23} & 59.1 & 74.3 & 84.8 & 22.6 & 40.7 & 58.0 & 79.2 \\
                                & PRISM \cite{cai2024prism}~\tiny{ACMMM'24} & {60.0} & {74.9} & {85.1} & 23.9 & 41.8 & 58.9 & - \\
                                & \graycell IMD (\textbf{Ours}) & \graycell \textbf{61.2} & \graycell \textbf{76.0} & \graycell \textbf{85.8} & \graycell \textbf{29.8} & \graycell\textbf{48.3} & \graycell\textbf{64.2} & \graycell \textbf{88.7} \\
                           \bottomrule
\end{tabular}
}
\label{tab-megadepth}
\vspace{-0.3cm}
\end{table*}

\subsection{Supervision}
\label{sec3.6}
The entire pipeline is trained in an end-to-end manner, with separate supervision applied to the coarse and refinement matching modules. 

\paragraph{Coarse-Level Matching Supervision.} 
The coarse-level loss is computed as the focal loss between the matching score matrix and the ground truth matches. The coarse ground truth matches, denoted as \(\{\mathcal{M}_c\}_{gt}\), with a total number of \(N\), are constructed by warping grid-level points from \(I_A\) to \(I_B\) using depth maps and image poses following \cite{sun2021loftr}. The resulting correlation score matrix \(\mathcal{S}\) in coarse matching is supervised by minimizing the log-likelihood loss over the locations in \(\{\mathcal{M}_c\}_{gt}\):
\begin{equation}
\mathcal{L}_c = -\frac{1}{N} {\textstyle \sum_{(\tilde{i},\tilde{j}) \in \left \{ \mathcal{M}_c \right \} _{gt} }^{}}  \log \mathcal{S}(\tilde{i  }, \tilde{j}).
\end{equation}

\paragraph{Fine-Level Matching Supervision.}

We train the fine-level matching module by supervising two losses independently. The first fine loss, \( \mathcal{L}_{f1} \), aims to minimize the log-likelihood loss of each fine local score matrix \( \mathcal{S}_l \) based on the pixel-level ground truth fine matches, similar to the coarse loss. The second is optimized using \( \mathcal{L}_{f2} \), which computes the \( \ell_2 \) loss between the final subpixel matches \( \{\mathcal{M}_f\} \) and the ground truth fine matches \( \{\mathcal{M}_f\}_{gt} \).

The total loss is computed as a weighted sum of all supervision terms: $\mathcal{L} = \mathcal{L}_c + \alpha \mathcal{L}_{f1} + \beta \mathcal{L}_{f2}$.

\section{Experiments}

\subsection{Implementation Details}
Following DIFT \cite{tang2023emergent}, we employ Stable Diffusion (SD) 2-1 \cite{rombach2022high} pretrained on LAION \cite{schuhmann2022laion} dataset as the diffusion model. We utilize the upsampling block index $n$ = 2 of the UNet so the feature map size is 1/8 of the input and the dimension is 640, and we set the time step used for the diffusion process to $t$ = 0. We choose CLIP \cite{radford2021learning} as the image encoder $\mathcal{V}$ in CIPM. 
We adopt the ResNet \cite{he2016deep} to extract fine image features. The self- and cross-attention are interleaved for $N$ = 2 times to transform coarse features. \(\tau\) is chosen to 0.2. Our model is trained on the MegaDepth dataset~\cite{li2018megadepth}, a large-scale outdoor dataset. Following the protocol in~\cite{wang2024efficient}, the test scenes are strictly separated from the training data. The weights $\alpha$ and $\beta$ in the loss function are set to $1.0$ and $0.25$, respectively. We employ the AdamW optimizer with an initial learning rate of $4\times10^{-3}$. The training process takes 30 epochs on $8$ NVIDIA 3090 GPUs. Both the coarse and fine stages are trained jointly from scratch. The model trained on MegaDepth is subsequently evaluated across all datasets to showcase its generalization capability.

\subsection{Evaluation Dataset for Multi-Instances}

\PAR{Datasets.} During the evaluation of IMD on conventional datasets for matching tasks, we observed that most existing datasets primarily have only a single instance per object category. For example, widely used benchmarks like MegeDepth \cite{li2018megadepth} and ScanNet \cite{dai2017scannet} focus on individual landmarks such as buildings, with each category typically representing only one instance. These characteristics make it easier for global semantic-based methods to get accurate matching results, as there are often no objects from the same category but different instances need to distinguish within or across images. As a result, comparing instance-based features against current methods on these benchmarks tends to yield similar results, which does not fully showcase the advantages of instance-based approaches.

To this end, we propose a new benchmark: Image Multi-Instance Matching (IMIM). IMIM is built using the BURST \cite{athar2023burst} dataset designed for Segemntation and Tracking inspired by \cite{samuel2024s}. This dataset consists of videos with pixel-level segmentation masks for each unique object track across different object categories. Specifically, we select videos that include at least two instances of the same object class and pick two frames from videos every time, assigning one as the source image and another used for the target image. After filtering, we obtain a total of 100 pairs images from 50 videos across 10 object categories.

\PAR{Evaluation Protocol.} 
Ground-truth masks are used to crop instances from source and target images. For every match results ($i$, $j$), we calculate the number $N$ of matching pairs that satisfied $i$ located in the source image instance mask. We report the percentage of $M/N$ that $M$ is the number of matching pairs which satisfied $i$ located in the source image instance mask, while $j$ located in the target image instance mask.

\PAR{Results.} Table \ref{tab-megadepth} presents the results in IMIM. IMD demonstrates superior performance over methods designed specifically for matching task and zero-shot models including CLIP, DINOv2, DIFT with diffusion, demonstrating our approach's reliable ability when matching the multi instances in the target image. 
Notably, we observe significant discrepancies between DINOv2 and DIFT. This due to DINOv2 ignoring the information not related to semantics. Qualitative comparisons are shown in Figure \ref{fig-m}, demonstrating that our approach consistently and accurately matches the correct instance in the target image.
Our experimental results validate the effectiveness of the proposed approach in significantly mitigating the misalignment issue.

\begin{table}[t]
    \caption{\textbf{Results of Homography Estimation on Hpatches Dataset.}} 
    \vspace{-0.3cm}
    \centering
    \resizebox{1.0\columnwidth}{!}{
    \setlength\tabcolsep{10pt} 
    \begin{tabular}{clccc} 
    \toprule
    \multirow{2}{*}{\textbf{Categeory}} & \multirow{2}{*}{\textbf{Method}} & \multicolumn{3}{c}{\textbf{Homography est. AUC}} \\
    \cmidrule(lr){3-5}
        & & @3px &  @5px & 10px \\
    \midrule
    \multirow{3}{*}{Sparse} & R2D2 \cite{revaud2019r2d2}+NN~\tiny{NeurIPS'19}  & 50.6 & 63.9 & 76.8 \\
    & OmniGlue \cite{jiang2024omniglue}~\tiny{CVPR'24} & 55.3 & 69.0 & 82.5 \\
    & SP \cite{detone2018superpoint}+SG \cite{sarlin2020superglue}~\tiny{CVPR'20} & 53.9 & 68.3 & 81.7 \\
    \hline
    \multirow{7}{*}{Semi-Dense} & LoFTR \cite{sun2021loftr}~\tiny{CVPR'21} & 65.9 & 75.6 & 84.6 \\
    & Effcient LoFTR \cite{wang2024efficient}~\tiny{CVPR'24} & 66.5 & 76.4 & 85.5 \\
    & HomoMatcher \cite{wang2024homomatcher}~\tiny{AAAI'25}  & 70.2 & 79.6 & 87.8 \\
    & SRMatcher \cite{liu2024semantic}~\tiny{ACMMM'24}  & 71.2 & 79.3 & 87.0 \\
    & CasMTR \cite{cao2023improving}~\tiny{ICCV'23}  & 71.4 & 80.2 & 87.9 \\
    & PRISM \cite{cai2024prism}~\tiny{ACMMM'24}  & {71.9} & {80.4} & {88.3} \\
    & \graycell IMD (Ours)  & \graycell \textbf{73.9} & \graycell \textbf{82.0} & \graycell \textbf{88.9} \\
    
    \bottomrule
    \end{tabular}
    }
    \label{tab-hpatches}
    \vspace{-0.6 cm}
\end{table}

\subsection{Relative Pose Estimation}

\PAR{Datasets.} We utilize the outdoor MegaDepth ~\cite{li2018megadepth} dataset and the indoor ScanNet \cite{dai2017scannet} dataset to evaluate relative pose estimation, showcasing the effectiveness of our approach. 

MegaDepth is an extensive outdoor dataset comprising 1 million images across 196 scenes, reconstructed using COLMAP \cite{schonberger2016structure}. The key challenges in this dataset include significant viewpoint and illumination variations, along with repetitive patterns. For evaluation, we adopt the test split from \cite{sun2021loftr}, using 1500 sampled pairs from the "Sacre Coeur" and "St. Peter’s Square" scenes. All images are resized such that the longest edge equals 1152 pixels.

ScanNet consists of monocular sequences with ground truth annotations and presents challenges due to wide baselines and textureless regions. For evaluation, we use the test pairs sampled in \cite{sun2021loftr}, resizing all images to 640 × 480 pixels for all methods.

\PAR{Evaluation Protocol.} Following \cite{sun2021loftr,wang2024efficient}, the recovered relative poses from the matches are evaluated to assess matching accuracy. The pose error is defined as the maximum of the angular errors in rotation and translation. We evaluate the performance by reporting the AUC of pose error at thresholds of 5\degree, 10\degree, and 20\degree.

\begin{figure*}[tp]
    \centering
    \includegraphics[width=1\linewidth]{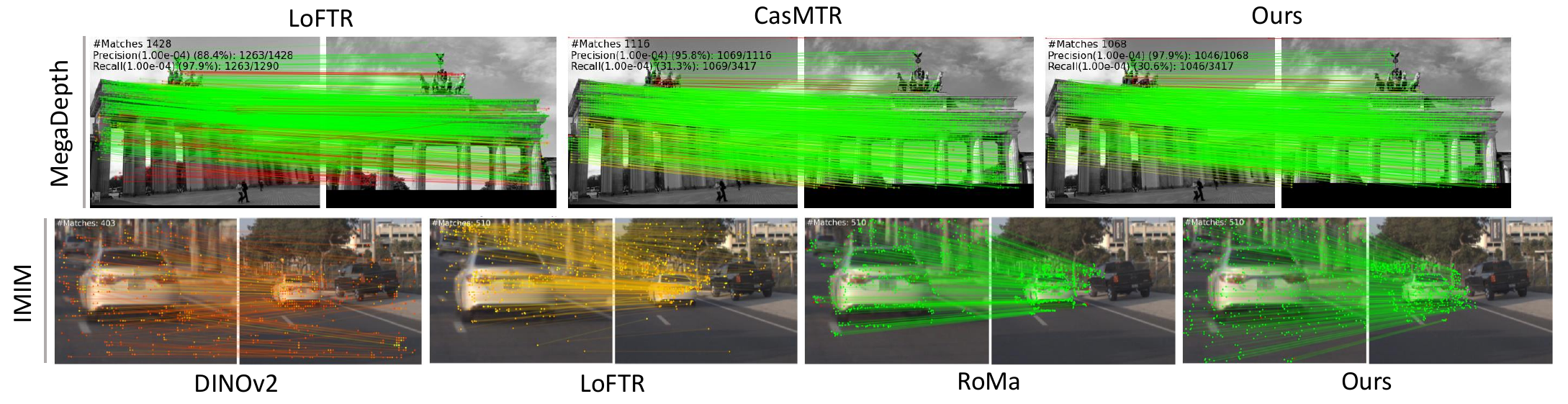}
    \vspace{-0.6 cm}
    \caption{\textbf{Qualitative comparison} of outdoor MegaDepth and IMIM datasets compared with DINOv2 \cite{oquab2023dinov2}, LoFTR \cite{sun2021loftr}, CasMTR \cite{cao2023improving}, RoMa \cite{edstedt2024roma} and ours IMD.  Green indicates the correctly matching results. More visualizations are provided in the Suppl.
     }
    \vspace{-0.4 cm}
    \label{fig-m}
\end{figure*}

\PAR{Results.} We report the accuracy comparison between IMD and other methods in Table \ref{tab-megadepth}. Qualitative comparisons are shown in Figure \ref{fig-m}.
Our proposed IMD sets a new state-of-the-art performance across all evaluation metrics on outdoor benchmarks.  
Benefiting from the use of pre-trained diffusion backbone, IMD notably improves by 24.6\% in AUC@5\degree on ScanNet compared to the best model trained on MegaDepth PRISM, showing impressive generalization capability.
We have the extended version of quantitative results (e.g. dense methods) and more analysis in the Suppl.

\begin{table}[t]
\caption{\textbf{Results of Visual Localization on InLoc Dataset.}} 
\vspace{-0.3cm}
\centering
\resizebox{1.0\columnwidth}{!}{
\setlength\tabcolsep{25pt} 
\begin{tabular}{ccc} 
\toprule
\multirow{2}{*}{\textbf{Method}}         & DUC1 & DUC2\\ 
\cmidrule(lr){2-3}
    & \multicolumn{2}{c}{(0.25m,2$\degree$)/(0.5m,5$\degree$)/(1.0m,10$\degree$)} \\ 
\midrule
SP \cite{detone2018superpoint}+SG \cite{sarlin2020superglue}~\tiny{CVPR'20} & 49.0 / 68.7 / 80.8 & 53.4 / 77.1 / 82.4 \\
LoFTR \cite{sun2021loftr}~\tiny{CVPR'21} & 47.5 / 72.2 / 84.8 & 54.2 / 74.8 / 85.5 \\
CasMTR \cite{cao2023improving}~\tiny{ICCV'23} & 53.5 / 76.8 / 85.4 & 51.9 / 70.2 / 83.2 \\ 
AspanFormer \cite{chen2022aspanformer}~\tiny{ECCV'22} & 51.5 / 73.7 / 86.0 & 55.0 / 74.0 / 81.7\\
PRISM \cite{cai2024prism}~\tiny{ACMMM'24} & 53.0 / \textbf{77.8} / 87.9 & 54.2 / 72.5 / 83.2 \\
\graycell IMD (\textbf{Ours}) & \graycell \textbf{54.6} / 77.3 / \textbf{88.2} & \graycell \textbf{60.3} / \textbf{82.7} / \textbf{88.5} \\
\bottomrule
\end{tabular}
}
\vspace{-0.5cm} 
\label{tab-inloc}
\end{table}

\subsection{Homography Estimation}
\PAR{Dataset.} Homography is crucial in two-view geometry, facilitating the transformation of perspectives between two images capturing the same scene. We evaluate on the widely used HPatches dataset \cite{balntas2017hpatches}, which includes 57 sequences with significant illumination variations and 59 sequences exhibiting significant viewpoint changes.

\PAR{Evaluation Protocol.} We present the area under the cumulative curve (AUC) for corner errors at thresholds of 3, 5, and 10 pixels. OpenCV’s RANSAC method is used for robust estimation. Following LoFTR \cite{sun2021loftr}, all test images are resized so that their shorter dimension equals 480 pixels.

\PAR{Results.} Table \ref{tab-hpatches} demonstrates that IMD significantly outperforms sparse and semi-dense methods. 
For the semi-dense methods, our method notably improves by 2.7\% in @3px compared to the best semi-dense methods. 
We attribute this to the effectiveness of guidance from the diffusion model for generalization and the proposed image interaction module for accuracy improvement.

\subsection{Visual Localization}

\PAR{Datasets and Evaluation Protocols.}
We evaluate the performance of IMD on the widely-used datasets InLoc \cite{taira2018inloc}. The InLoc dataset includes 9,972 RGBD images with geometric registration and 329 query images with verified poses, posing challenges in textureless or repetitive environments. For testing, we focus on DUC1 and DUC2 following \cite{sun2021loftr}. 
The candidate image pairs are identified using the pre-trained HLoc \cite{sarlin2019coarse} system following \cite{cai2024prism}.
The results of Aachen Day-Night v1.1 \cite{sattler2018benchmarking} are provided in Suppl.
\PAR{Results.} Table \ref{tab-inloc} presents the result on the InLoc dataset, IMD surpasses all methods on DUC2 and matches the performance of top-tier approaches on DUC1. 
IMD exhibits robust performances, underscoring the adaptability of our method in diverse task environments.

\subsection{Ablation Studies}
\label{ablation}

\PAR{Visual Representations.} We compare the internal representations of diffusion models with those of other pre-trained contrastive learning objective and discriminative objective models. 
We observe that IMD outperforms the other models on both datasets. 
This highlights that internal representations of diffusion models are indeed more efficient for matching tasks, especially in multi-instance scenarios. More comparison results shown in Suppl.

\PAR{Cross-image Interaction Prompt Module.} 
We explore the influence of interaction in CIPM on matching accuracy. Specifically, we assess the additional approaches: an empty string and the individual image prompt in Table \ref{tab-ablation}.  
These evaluations are regarded as does not presuppose any information interaction.  
These setups yield lower accuracy, further highlighting the benefits of the interaction between image pairs, and facilitates more precise capturing of cross-image relationships.
When we replace the cross-attention with a concatenation operation to execute interaction between images which means both images share the same prompt, there is a decline in performance in row 5.   

\PAR{Different Time Steps.} We also investigate which diffusion step(s) are most effective for feature extraction. As the value of $t$ increases, the amount of noise distortion added to the input image also increases. In Stable Diffusion \cite{rombach2022high}, there are a total of 1000 time steps. From the row 6 of Table \ref{tab-ablation}, we observe that all metrics decrease as $t$ increases, with the best results achieved at $t$ = 0 (ours). We conduct more ablation studies about timesteps and block-index in Suppl.

\begin{table}[t]
    \caption{\textbf{Results of Ablation Studies.}
    }
    \vspace{-0.3cm}
    \centering
    \resizebox{1.0\columnwidth}{!}{
    \begin{tabular}{lcccc} 
    \toprule
    \multirow{2}{*}{Method}         & \multicolumn{3}{c}{\textbf{MegaDepth dataset}}  & \multirow{2}{*}{\textbf{IMIM}}\\ 
    \cmidrule(lr){2-4}
        & AUC@5\degree       & AUC@10\degree       & ACU@20\degree \\ 
    \midrule
    
    1) Replace SD to Swin (B) \cite{liu2021swin} & 57.5 & 73.2 & 83.6 & 74.0\\ 
    2) Replace SD to DINOv2 (B) \cite{he2022masked} & 57.8 & 73.5 & 83.7 & 75.5\\
    3) Empty string  & 58.4 & 73.8 & 84.1 & 84.0\\
    4) Individual prompt in CIPM & 59.6 & 74.3 & 84.5 &85.2\\
    5) w/o cross attention & 60.7 & 75.0 & 85.1 &87.4\\
    6) time step T = 100 & 60.9 & 75.7 & 85.8 &88.1\\
    \textbf{Ours Full} & \textbf{61.2} & \textbf{76.0} & \textbf{85.8} &\textbf{88.7}\\
    \bottomrule
    \end{tabular}
    }
    \label{tab-ablation}
\end{table}

\subsection{Understanding IMD}
\begin{figure}[h]
    \centering
    \includegraphics[width=0.75\linewidth]{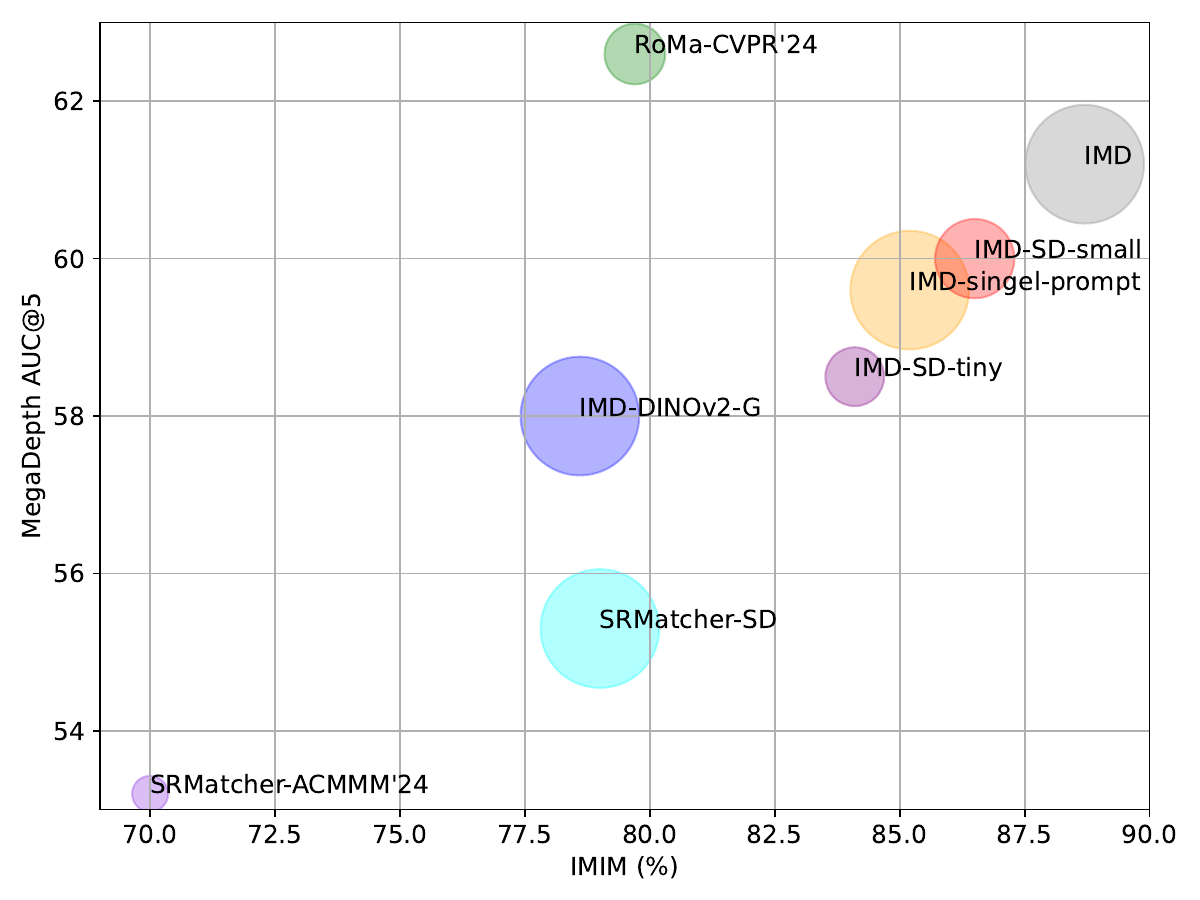}
    \vspace{-0.4 cm}
    \caption{\textbf{Performance comparison} and the circle size indicating parameter count.}
    \vspace{-0.5 cm}
    \label{fig-4}
\end{figure}

To mitigate potential bias, that the improvement of the model mostly achieved from increased parameters in SD 2-1 (1,300M), we compare with different backbone: (1) DINOv2(G) (1,100M) of similar scale; (2) SD-tiny and SD-small with CIPM, having 75\% and 55\% fewer parameters respectively. We further compare with RoMa \cite{edstedt2024roma} (dense) and SRMatcher \cite{liu2024semantic} (semi-dense) using DINOv2 and its SD 2-1 variant.
Figure \ref{fig-4} shows results on MegaDepth and IMIM. All SD-based models achieve strong IMIM performance, indicating capacity alone does not address multi-instance problems. IMD-single-prompt results validate cross-image interaction effectiveness. 
Discussion on feature behavior analysis and training protocol ablations are detailed in Suppl.

\section{Conclusion}
In this work, we present IMD, a novel framework designed to address the misalignment between vision foundation models and feature matching tasks. We identify two challenges for addressing the misalignment: 1) foundation model properties' impact on matching; 2) the disconnect between foundation models and feature matching. We also introduce a new benchmark specifically designed to assess misalignment mitigation. Extensive experiments show that IMD achieves major gains and effectively alleviates misalignment. 
We expect our work will inspire future research on how to better utilize vision foundation models for downstream tasks, while also contributing to the broader utilization of these foundational models across diverse tasks.

\section{Acknowledgment}
This work was supported in part by the National Natural Science Foundation of China under Grants 62088102 and National Science and Technology Major Project (No. 2023ZD0121300).

{
    \small
    \bibliographystyle{ieeenat_fullname}
    \bibliography{main}
}

\end{document}